# Investigation of Warrior Robots Behavior by Using Evolutionary Algorithms

Shahriar Sharifi Borojerdi[1] ,Mehdi Karimi[2] ,Ehsan Amiri[3]

***Abstract***—In this study, we review robots behavior especially warrior robots by using evolutionary algorithms. This kind of algorithms is inspired by nature that causes robots behaviors get resemble to collective behavior. Collective behavior of creatures such as bees was shown that do some functions which depended on interaction and cooperation would need to a well-organized system so that all creatures within it carry out their duty, very well. For robots which do not have any intelligence, we can define an algorithm and show the results by a simple simulation.

## I. INTRODUCTION

IN our periphery world, the entire live organisms composed of organized structures. Structures that are placed within creature's ventricle from the direction of universe creator. All these creatures composed of life basic block called "cell". Aforesaid rules exist in all the organisms in form of coded gens. By connecting these gens, was made lengthy strings called "Chromosomes". Each gene represents one of organism characteristics.

Here, it had taken advantage of ad hoc network to transmit data. Robots be scattered in geography situations and the interaction between those which had situated far from each other monitored by a multi-hop tracker.

Genetic algorithms (GA) use principles of Darwinian natural selection for finding optimum formula to predict or adopt a pattern. Genetic algorithms, as one of solutions had known among common approaches in artificial intelligence. In fact, by this way, we can take faster speed in problem state space for finding possible answers, it means we can do not expand all states to take the related answers.GA is a specific set of evolutionary algorithms ,it applies techniques which was stimulated by biology such as Selection, Mutation, Iteration, Crossover.

In general, evolutionary algorithms applied without supporting via other definitive algorithms, in addition it had successfully used for problem solving. In this study, we adopted a dynamic method with problem at different interval time.

A robot can moves into one out of six neighbor's cells, but its speed and direction is depends on other robots speed and directions. Movement method, at first was using evolutionary algorithm randomly then its movement got stable and its speed and directions calculated by evolution algorithm. There are numerous applications for topology control algorithms which depended on evolution algorithm, such as; maintaining facilities, emerging reference field to reach to a local research….[1].

At all these applications, there is data distribution problem all over the geographic site so that

1. Geographic site may be change in small limit 2. Agent numbers maybe change (dynamically get loss or more) 3.Agents cannot access to guiding map or GPS ,but be able to achieve limited data from their neighborhoods 4.Agents entered a site via an importing point (in comparison with random state or other distribution form are more realistic).Evolutionary algorithm especially GA ,ants and bees colonies are the good selection for dissolving these kind of problems, so that they can move toward a better solution and change with respect to environment situations. According to our investigation, application of evolutionary algorithm in these kinds of problems would be a more modern application.

In the second section on presented evolutionary algorithm and in the third section about robots behaviors by using algorithms and at last the results were presented.

## II. EVOLUTIONARY ALGORITHM

The diagram below is a general diagram concerning intelligent methods which among them, Evolutionary algorithm acts better for searching issue, however ,in this study did not perform any comparisons between them.

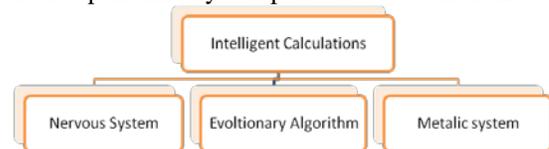
Fig. 1 Intelligent Methods

What is the relation between natural selection with artificial intelligence? The main reason of applied intelligent methods in artificial intelligence would be finding the optimal engineering issues. For example, how do we design an engine to achieve the best output or how do we mobile the robotics' arm to pass the shortest route?(notice in case of barrier existence, finding the shortest route would not be as simple as drawing a straight line from origin to destination),all them are optimization problems.

---

[1]Azad University Branch,Ghazvin,Iran sharifi.shahriar@gmail.com
[2]Azad University Branch,Ghazvin,Iran karimi_mh@yahoo.com
[3]Azad University Branch,Ghazvin,Iran e.e.amiri@gmail.com
.

Evolutionary algorithms under discussion could be divided to subsets of some methods such as Genetic algorithm, Ants' colony and bees' colony. However, the algorithms will discuss in this section, somehow are inspired by nature.

Genetic algorithms use principles of Darwinian natural selection to find optimum formula in order to predict or adopt the pattern. This is natural selection rule where just some species to breed by having the best characteristics, so those who have not these kind of characteristic become extinct gradually and over the time. Briefly, it was mentioned that genetic algorithms (GA) is a programming technique that it applies evolutionary genetic as a problem solving pattern. The problem that should be solved is input and solutions are coded based on a pattern and fitness function assesses any candidate solution so that most of them are selected randomly. In genertal, solution has shown as pair form 0 and 1 but there are other ways to display. Evolution was started by an entire randomized creatures set and it would repeat in the next generations. In each generation was selected the most suitable ones not the better ones. [4]

Using ants' algorithm had represented first time in 1991 by Dorigo and his colleagues to solve the complicated optimal issues so that it had drawn their attention then in 1996 and 1997 ants' colony algorithm was represented by Dorigo and his colleagues. Ants are a sort of insects that live as colony and collectively forms; as they haven't eyesight, they find the nest route to food source by secretion of" pheromone"[4] as a chemical material. Whenever an ant exits nest, in its route until reach to food source, it leaves trace of a pheromone so that the other ants follow the marked route by scenting and they are guided to food source. Simultaneously, these ants induce more concentration in pheromone by leaving pheromone in those routes. So, pheromone concentration is higher in routes which ants pass more. [5]

Bees 'organizational system is established by a series of insects' simple foreign principles. Each bee prefers to follow the pervious bees' route till it looks for new flower, itself. Each hive has known place called "dancing saloon" where bees together doing specific motion to convince co- hive bees to follow and select their routes in order to reach flowers. If a bee decided to collect nectar so by selecting co-hive bee, it follows the pervious route to reaches flowers. After reaching to flowers and collecting nectars, bees are able to perform these following tasks:

A. Release the food source and again looking for a dancer (leader) bee to find a new food source.

B. Searching by itself for new food source.

C. It starts to dance in the hive in order to bring and motivate the new bees to follow it.

Search process is composed of successive reiterations. The first reiteration is finished whereas the first bee had presented its sub-basic solution to solve the main problem.

The best sub-basic solution had selected during the first reiteration and then will start the second reiteration. In the second reiteration, artificial bees are starting work to find new solution for smaller problems and at the end of each reiteration would be one or some solutions.

## III. BEHAVIOR STATE OF WARRIOR ROBOTS BY USING EVOLUTIONARY ALGORITHMS

The important issue concerning robots interaction is this issue to share data and knowledge. In this field, there are numerous methods; here, we review these methods. These methods should be able to share the data and knowledge all over the geographic sites and among the other robots in order to data and knowledge would be transmitted better. For transmitting these data used evolutionary algorithms were inspired by nature. Here we take advantages of ad hoc networks and by using evolutionary algorithms, we represent analytical model to transmit the data among the warrior robots that is performed by simulated software with C# language.

### A. Robots behavior by using Genetic algorithms

Here, we start our work with introducing transmitting data as known chromosomes. Genetic algorithm controls robots actions- to make interaction among them- by investigation on new existent chromosomes and their specific functions. A point was shown concerning this algorithm, robot's tracking to reach the target site was excellent so that algorithm was able to guide the robot to target site in numerous periods ,but group transmission of robots have some problems that it is referred to algorithms nature.

During to use GA, each chromosome contains speed, direction, degree and situation of agent knot. Each robot has a kinetic agent that genetic algorithm decides about its direction and speed. This algorithm act at selection part, differently, this difference is referred to motion speed and direction. Each robot could be situated in one home so that the home assumed as hexagonal. Because of this reason, we assumed it as hexagonal to be able to make interaction and interconnection as far as possible .Other geometric forms have faults, for example square -with at least sides -cannot establish that assumption in the mind, as well. The knots situated at center, can move toward one out of six directions around it.

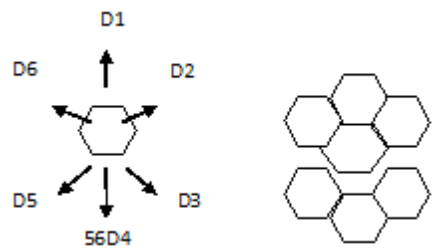

Fig 2. Shape 1- Place situation of a robot

If the hexagonal numbers was rose up as large as our problem space, we see each slipknot be able to move toward all six directions around it.

Sometimes it would be possible that nodes close to marginal sites which in these cases we assign those sites as inaccessible for nodes, because we want to extend the nodes in our sites so they have to select another region for the next movement .

Each robot placed in one of these six cells and to make interaction with other robots, it transmits and sends its data to one of neighboring cells based on its speed and its movement direction. Neighboring robot processes the data after receiving them then it makes decision for movement and attack.

Shape below is a kind of robots movement by using genetic algorithm that was designed to attack to a target.

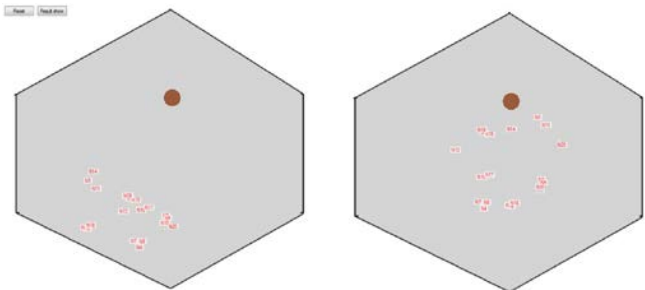

Fig. 3. Shape2- Investigation by using genetic algorithm

Here, it was assigned 20 robots that they perform movement action toward the target by sending and transmitting data to each others. Here, primary movement is based on randomized movement. After starting the movements it was shown that some robots move more toward the target and some others move lesser toward the target, it would be because of its primary definition in the data transmitting.

### B. Robots behavior by using ants' colony

Here, we start our work by introducing each robot as an ant. In this method, data transmitting is defined depend on speed rate and closeness of other robots to target. Here, existent pheromone in fact acts as transmitting data for robots so that whatever more robots to close to a target, more data to transmit and the other robots move based on received data rates from neighboring robots.

Here, according to shape1, we assumed the environment as an hexagonal so that each robot can move toward neighboring sides or be able to send the data to those regions. Shape below is a kind of robots movement action by using ants' colony.

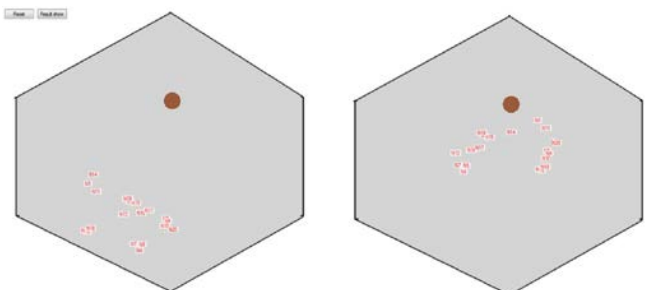

Fig. 4. Shape 3- Investigation by ants' colony

Here, a robot movement is done in chain-like form that data is received from neighboring robots and movements is done toward target. As it had shown, the robots movement divided to two categorizations, it was because of this reason that here, two robots more transmitted the data versus the others.

### C. Robots behavior by using bees' colony

Here, we start our work by introducing each robot as a worker bee and define a robot as a dancer (leader) robot. In this method, Transmitting data defined depend on dancing rate of leader robot. Robot- N1 tackles dancer robot duty and defines the other robots direction and movements.

Here, according to above shape we assumed environment as a hexagonal so that each robot can move toward its neighboring sides or transmit the data to those regions. Shape below is a kind of robots movement action by using bees' colony.

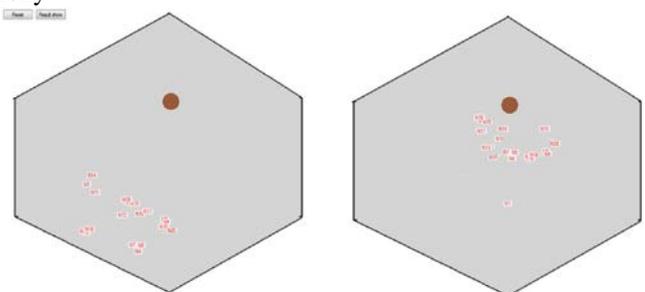

Fig. 5. Shape 4- Investigation by bees' colony

Here, the other robots take order from the robot N1 and as you seen; they occupied more regions of space. If the robot N1 to distinct, we should demonstrate a mechanism which defines the other best robot.

### IV. Conclusion

In this study robots interconnection is investigated based on interaction, movement and functions on a target that it was done by using genetic evolutionary algorithms, ants' colony and bees' colony.

Based on this investigation, robots moved toward the target and it was defined by defining and optimization of these algorithms, it is possible to achieve satisfying results.